\documentclass[12pt,a4paper]{article}
\usepackage[T1]{fontenc}
\usepackage{lmodern}
\usepackage[english]{babel}
\usepackage{graphicx}
\usepackage{caption}
\usepackage{amsmath}
\usepackage[toc]{appendix}
\usepackage{amsfonts}
\usepackage[numbers]{natbib}
\usepackage{xcolor} % Package for in-text citation colorization
\usepackage{CJKutf8}
\usepackage{booktabs}
\usepackage{colortbl}
\usepackage{array}
\usepackage{makecell}
\usepackage{authblk}
\usepackage{listings}
\usepackage{tcolorbox}
\usepackage[colorlinks, citecolor=blue]{hyperref} % Ensure proper font encoding
\usepackage{breakurl}

\title{\LARGE{Automatic Essay Multi-dimensional Scoring with Fine-tuning and Multiple Regression}}
\date{\vspace{10ex}}
\author[1]{\footnotesize Kun Sun\thanks{\texttt{email: kun.sun@uni-tuebingen.de}}}
\author[2]{\footnotesize Rong Wang \thanks{\texttt{email: rongw.de@gmail.com}}}
\affil[1]{\footnotesize Department of Linguistics, University of Tübingen, Germany}
\affil[2]{\footnotesize Institute of Natural Language Processing, Stuttgart University, Stuttgart, Germany}

\begin{document}
	
	\maketitle
	
	\vspace{0.5cm}
	
	\begin{abstract}
		Automated essay scoring (AES) involves predicting a score that reflects the writing quality of an essay. Most existing AES systems produce only a single overall score. However, users and L2 learners expect scores across different dimensions (e.g., vocabulary, grammar, coherence) for English essays in real-world applications. To address this need, we have developed two models that automatically score English essays across multiple dimensions by employing fine-tuning and other strategies on two large datasets. The results demonstrate that our systems achieve impressive performance in evaluation using three criteria: precision, F1 score, and Quadratic Weighted Kappa. Furthermore, our system outperforms existing methods in overall scoring.
	\end{abstract}

\small{{\bf Keywords:} {multiple scoring, classifier, mutiple regression, fine-tuning, LLMs}

\clearpage

\section{Introduction}
% represents a significant advancement in the application of natural language processing (NLP) within the educational sector.  
Automated Essay Scoring (AES) could automatically evaluate the proficiency of written essays. When AES works with high effectiveness, this technology not only saves educators substantial time that would otherwise be spent on manual essay grading but also provides students with immediate and free feedback. Moreover, AES systems offer more consistent and impartial assessments compared to human evaluators. More importantly, AES could be commercially applied in computer-aided language learning market.

Over the past half-century, a diverse array of methodologies has been proposed to tackle the challenges of AES. These methodologies range from learning from rule-based features to machine learning methods \cite{ke2019automated}; \cite{ramesh2022automated}. The recent development is to leverage neural approaches, including pre-trained language models \cite{dong2017attention}, \cite{yang2020enhancing}. After the revolution of transformer-based language models and large language models (LLMs), the AES work based on LLMs has achieved STOA performance. This process is quite similar to other tasks in NLP. The primary objective of most AES research is to predict an overall holistic score that aligns closely with human judgment. Additionally, other studies have focused on providing detailed feedback by estimating quality scores across multiple traits of an essay.

We summarize the past work on AES. First, there are several approaches to AES: regression model \cite{taghipour2016neural} where the goal is to predict the score of an essay; \cite{cummins2016constrained}; \cite{dong2017attention}, classifier model, where the goal is to classify an essay as belonging to one of a small number of classes (e.g., low, medium, or high, as in the TOEFL11 corpus) \cite{nguyen2018argument}. \cite{vajjala2018automated}, \cite{ke2019give} and ranking model \cite{yannakoudakis2011new}, \cite{chen2013automated}, \cite{dong2016automatic}, where the goal is to rank two or more essays based on their quality. Second, the vast majority of existing AES systems were developed for holistic scoring. Dimension-specific scoring did not start until 2004. So far, several dimensions of quality have been examined, such as, organization \cite{persing2010modeling}, argument persuasiveness \cite{ke2018learning}, coherence \cite{somasundaran2014lexical}. However, these systems have not been developed based on LLMs.  Third, the datasets were mostly relying on two: ASAP  (Automated Student Assessment Prize) and TOEFL11, and both merely include the information holistic score.

Recent advancements in AES have been driven by the use of transformer-based language models, which achieve state-of-the-art performance through combined regression and ranking optimization techniques \cite{yang2020enhancing}, \cite{xie2022automated}, \cite{jiang2023improving}. Furthermore, with proper prompting strategies, general-purpose language models like ChatGPT and LLaMA can also facilitate AES for small-scale essays \cite{lee2024applying}, \cite{mansour2024can}, \cite{latif2024fine}. However, when it comes to processing a large volume of essays, specialized APIs and appropriate prompts are required for efficient handling. Our focus is on these specific AES that enables spontaneous and comprehensive processing of numerous essays (e.g., \cite{wang2022use}, \cite{xie2022automated}).

Overall, AES stands as a pivotal innovation in educational technology, promising to enhance the efficiency and fairness of essay assessments. As NLP continues to evolve, AES systems are expected to become even more sophisticated, further bridging the gap between automated and human scoring. 
  
Despite the impressive results achieved by models designed for AES, several challenges and limitations persist. The following specifies these limitations.

 \underline{Model training on \texttt{BERT}}: A number of models for AES have been trained using BERT, a general-purpose language model. However, BERT might not be the optimal choice for directly enhancing AES effectiveness. Scoring essays could be fundamentally takne as a text classification task to some degree, and improvements can be made from an engineering perspective by fine-tuning existing models specialized for text classification. % Models used in sentiment analysis or emotion detection \cite{rodriguez2023review}, \cite{zhang2023sentiment}, for instance, have achieved impressive performance compared with other text classifier, and these existing model could be fine-tuned to enhance their ability for offering more precise scoring capabilities when adapted for AES.

\underline{Limitations of training datasets}: The training datasets for AES often rely heavily on the  ASAP essays, which are written by students in grades 7 and 10 in USA (i.e., junior secondary school students). These essays may not provide the necessary complexity and depth for robust AES training. Currently, there are superior datasets available that include essays from students in grades 10 to 12. These datasets offer a richer variety of language use and structure, making them more suitable for training advanced AES systems.

 \underline{Holistic scoring limitations}: Traditional AES systems typically provide a single holistic score for each essay. While this approach simplifies the scoring process, it fails to meet the nuanced expectations of real-world users. Ideally, an advanced AES system should deliver multiple scores across different dimensions such as vocabulary, grammar, coherence, and overall quality. This multi-dimensional evaluation would offer a more comprehensive and objective assessment of essays, providing more valuable feedback to second-language learners and researchers.

%\underline{Regression and ranking methods}: Although some models, such as the one proposed by Yang et al. (2020), combine regression and ranking techniques, they still produce only one overall score per essay. To address this, we propose developing a new AES system that can generate multiple scores across different dimensions using a multiple regression model. This approach would provide a more detailed and nuanced assessment of each essay.

%5) Understanding Prompts: Different essays have varying requirements regarding style, topic, and prompts. To enhance the ability of classifiers to understand and respond to these prompts, incorporating algorithms that specialize in contrastive learning could be beneficial. Contrastive learning can improve the model's sensitivity to prompt-specific nuances during fine-tuning, leading to more accurate and context-aware scoring.

In summary, while current AES models have achieved significant progress, there is room for improvement in several areas. By addressing these challenges, we can develop more effective and versatile AES systems that provide richer, more detailed feedback and are better suited to the diverse needs of educational contexts.

To overcome these limitations, we propose new strategies and utilize an enhanced dataset to train a more robust AES system capable of grading essays across multiple dimensions. This new system is termed as ``automatic essay multi-dimensional scoring'' (\textbf{AEMS}). This is the primary objective of this paper. The following sections detail the methods employed to train AEMS and present our experimental results.

\section{Methods}

\subsection{Technique route}
\label{tecrou}
As discussed in the introduction section, we adopted a novel approach to design a new multi-dimensional scoring system. %Leveraging the impressive performance of sentiment classifier models, 
We first selected several effective classifiers and fine-tuned them. Next, we aimed to implement multi-dimensional scoring, a process akin to multiple regression, for which we employed a multiple regression model during fine-tuning. Additionally, some essays include extra information, such as requirements, topics, and types. We applied contrastive learning to enable the models to better understand and incorporate this information, enhancing the accuracy of the scoring. The following details these techniques.

%\section*{Multi-Class Classification with BERT for Sentiment Analysis}

First, for multi-class classification using BERT or RoBERT or other BERT-based classifiers, we obtain the representation of the \texttt{[CLS]} token from the BERT model. This representation is then passed through a dense layer with a \texttt{softmax} activation to predict class probabilities.

Let $\text{BERT}(x)$ be the output representation of the [CLS] token for input $x$;  $W$ be the weight matrix of the dense layer; $b$ be the bias vector; $\hat{y}$ be the predicted probability vector for $K$ classes. The predicted logits $z$ are given by:
\[
z = W \cdot \text{BERT}(x) + b
\]

We then apply the \texttt{softmax} function to obtain the class probabilities:
\[
\hat{y}_i = \frac{e^{z_i}}{\sum_{j=1}^K e^{z_j}} \quad \text{for} \ i = 1, 2, \ldots, K
\]

%\section*{Multiple Regression Combined with BERT for Sentiment Analysis}

Second, to incorporate multiple regression, we add a separate regression head to the \texttt{[CLS]} token representation. Let $\beta$ be the regression weight vector; $\gamma$ be the regression bias; $y$ be the continuous target variable. The regression model is expressed as:
\[
y = \beta \cdot \text{BERT}(x) + \gamma
\]

%\section*{Combined Model and Loss Function}

The combined BERT-based model has two heads: one for classification and one for regression. The overall loss function combines cross-entropy loss for classification and mean squared error (MSE) loss for regression.

Let $\mathcal{L}_{\text{CE}}$ be the cross-entropy loss; $\mathcal{L}_{\text{MSE}}$ be the mean squared error loss; $\lambda$ be the weight balancing the two losses. The combined loss function is:
\[
\mathcal{L} = \mathcal{L}_{\text{CE}} + \lambda \mathcal{L}_{\text{MSE}}
\]

%Specifically:
%\begin{itemize}
%    \item \textbf{Cross-Entropy Loss:}
%    \[
%    \mathcal{L}_{\text{CE}} = - \sum_{i=1}^N \sum_{k=1}^K y_{i,k} \log(\hat{y}_{i,k})
%    \]
%    where $N$ is the number of samples, $y_{i,k}$ is the true label for sample $i$ and class $k$, and $\hat{y}_{i,k}$ is the predicted probability for sample $i$ and class $k$.
    
%    \item \textbf{Mean Squared Error Loss:}
%    \[
%    \mathcal{L}_{\text{MSE}} = \frac{1}{N} \sum_{i=1}^N (y_i - \hat{y}_i)^2
%    \]
%    where $y_i$ is the true continuous score for sample $i$ and $\hat{y}_i$ is the predicted continuous score for sample $i$.
%\end{itemize}

By combining these losses, the BERT-based model can be trained to perform multi-class classification and regression simultaneously. %, enhancing sentiment analysis capabilities. 
The contrastive learning allows to better learn the information on essay prompts. The formal description on this is seen in \textbf{Appendix A}.

Our primary approach involves fine-tuning existing text classification models and applying multiple regression methods to improve the effectiveness for multi-dimensional scoring, and enhancing their capabilities through contrastive learning to better understand essay prompts.  To develop a more practical AES for second language (L2) learners' essays, we employed a new dataset, the English Language Learner Insight, Proficiency, and Skills Evaluation Corpus (ELLIPSE) \cite{crossley2023english}, which contains 9,000 essays written by students in grades 8 to 12 in USA. Additionally, to ensure applicability in real-world scenarios, we trained our models on the official IELTS (International English Language Testing System) exam dataset %\footnote{The dataset is available at \burl{https://www.kaggle.com/datasets/mazlumi/ielts-writing-scored-essays-dataset}} in order to cross-validate with the first training dataset by using the same strategies.
The following sections specify our strategies and training datasets in detail.

\subsection{Fine-tuning existing Models}
Given that scoring is a classification task, we can fine-tune existing high-performance models to tailor them for the AES purpose. To validate our strategies, we selected two existing models for fine-tuning:

%\texttt{cardiffnlp/twitter-roberta-base-sentiment-latest}:
 %This RoBERTa-base model \cite{camacho2022tweetnlp}, 
% trained on approximately 124 million tweets from January 2018 to December 2021, has been fine-tuned for sentiment analysis using the TweetEval benchmark. It excels at understanding and categorizing sentiments in English-language tweets, providing enhanced accuracy for social media monitoring, opinion mining, and market research. Leveraging a vast dataset and the robust RoBERTa architecture, this model delivers precise sentiment classification, distinguishing between positive, negative, and neutral sentiments. Its adaptability to evolving language and trends on Twitter makes it a valuable tool for contemporary sentiment analysis.
The RoBERTa-base model is good at making classifications, and we can take advantage of this classifier to fine-tune. The RoBERTa works well in making fine-tuning.  \texttt{Distil\-BERT} is a smaller, more efficient language representation model pre-trained using knowledge distillation. % \cite{sanh2019distilbert}. 
This method reduces the BERT model size by 40\%, retains 97\% of its language understanding capabilities, and is 60\% faster. The model employs a triple loss function combining language modeling, distillation, and cosine-distance losses. Through proof-of-concept and on-device studies, DistilBERT has demonstrated its efficacy as a cost-effective solution for high-quality NLP tasks in constrained environments.

%\texttt{distilbert/distilbert-base-uncased-finetuned-sst-2-english}: 

In order to make these models suitable for AES, we fine-tune them to scoring essay quality across various dimensions such as syntax, vocabulary, and coherence. %By leveraging their strengths in sentiment and language understanding, 
We used multiple regression and contrastive learning aim to enhance their ability to assess and score essays more comprehensively. The tech details were described above. The tech roadmap is seen in Fig. \ref{road}. The following sections will detail the datasets used, and our fine-tuning process.

\begin{figure}
\centering
\includegraphics[width= 1\textwidth]{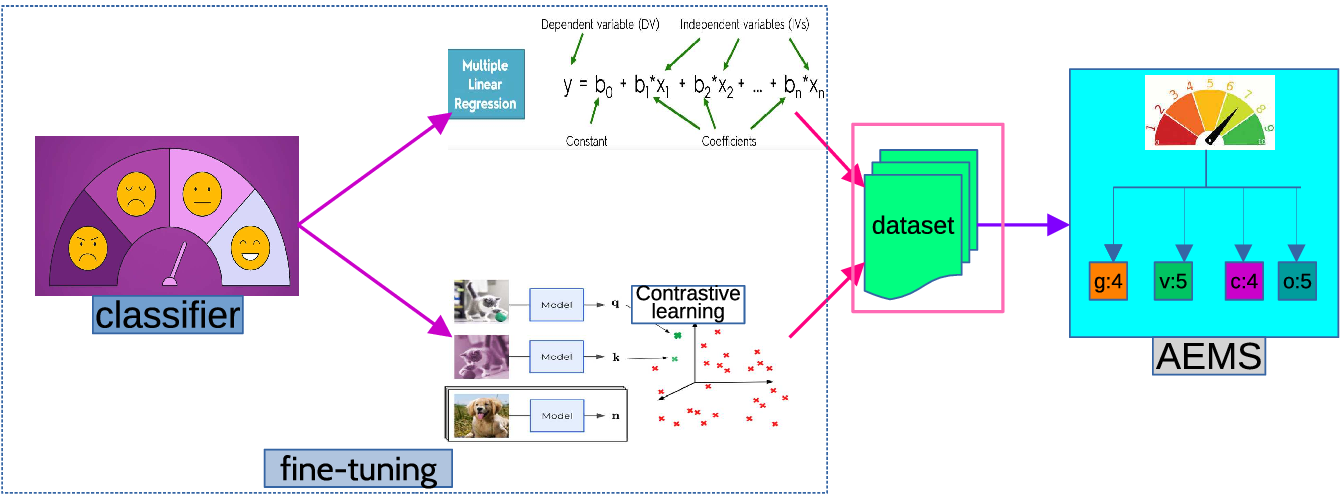}
\caption{The roadmap of developing AEMU in the present study}
\label{road}
\end{figure}

\subsection{Training datasets}
To support our research, we utilize two comprehensive datasets. The following details the two datasets for training and testing. 

 The ELLIPSE Corpus is a freely available resource containing approximately 9,000 writing samples from English Language Learners (ELL) \cite{crossley2023english}. These samples are scored for overall holistic language proficiency and analytic proficiency in various dimensions, including cohesion, syntax, vocabulary, phraseology, grammar, and conventions. Additionally, the corpus provides individual and demographic information about the writers, such as economic status, gender, grade level (8-12), and race/ethnicity. Developed to support research in corpus and NLP approaches, the ELLIPSE Corpus offers detailed language proficiency scores, aiding in the assessment of both overall and specific aspects of language proficiency. For our research, the training dataset consists of 8,100 essays, while the test dataset includes 900 essays.

The second dataset comprises essays from the International English Language Testing System (IELTS). This dataset includes the prompt, the essay itself, and scores across six dimensions: task achievement, coherence and cohesion, vocabulary, grammar, and an overall score. This dataset is particularly valuable as it reflects real-world scoring criteria used in high-stakes language assessments. For our purposes, the training dataset includes 14,500 essays, and the test dataset contains 2,000 essays.

By leveraging these datasets, we aim to fine-tune our models to provide detailed, multi-dimensional scoring of essays. The following sections will detail our methodologies for fine-tuning, the specific strategies we employed, and the results of our experiments. 

\subsection{Evaluation criteria}

 %Our approach involved two primary strategies to improve the model's scoring capabilities.

%The first strategy involved adopting a regression model. This approach treats scoring as a regression problem, enabling the model to predict continuous scores that align more closely with human assessments. The second strategy incorporated contrastive learning during the fine-tuning process. This method enhances the model's ability to generalize across various prompts, styles, and requirements.

%Our training datasets included diverse prompts. For instance, the ELLIPSE dataset features essays with different types, styles (expository and argumentative), and topics. The IELTS dataset includes essay titles and specific requirements from the exams. These variations in prompts significantly impact scoring, and contrastive learning helps the model adapt to these differences, improving its overall robustness.

Following these strategies, we employed either of the selected models and utilized comprehensive training datasets to fine-tune them into the AEMU systems. Upon completing the model training, we validated the model's effectiveness using the corresponding test datasets. The evaluation standards included the following metrics: \texttt{precision}, \texttt{F1 score}, and \texttt{Quadratic Weighted Kappa (QWK)} for each dimension in each test dataset. This means that each essay in the test dataset was evaluated across various dimensions—such as vocabulary and grammar—using all three criteria. For example, in the first test dataset, the dimension ``vocabulary'' was assessed based on precision, F1 score, and QWK. Similarly, the ``grammar'' dimension was also evaluated using these same three criteria. This comprehensive evaluation approach ensures that the models' performance is thoroughly assessed from multiple angles, providing a detailed understanding of their strengths and weaknesses in scoring different aspects of essay quality.

Overall, to facilitate an explicit comparison, we employed the same fine-tuning strategies on two different existing models— %``cardiffnlp/twitter-roberta-base-sentiment-latest'' and 11distilbert/distilbert-base-uncased-finetuned-sst-2-english''—and 
two training datasets (ELLIPSE and IELTS). This resulted in four distinct models. We also made comparison with the previous work. %By adopting these methods, we aim to demonstrate the effectiveness of our fine-tuning strategies and highlight the improvements in automatic essay scoring capabilities, offering a more unique, practical and comprehensive assessment of L2 English writings. %These models were then tested on the Automated Student Assessment Prize (ASAP) dataset, which has been used extensively in previous research.

\section{Results}

\subsection{Study 1}
Following the similar strategies to fine-tune and retrain the two existing models on the same two datasets, we obtained two AEMU models. The first one is termed \texttt{RoAEMS} (RoBERT-based Automatic Essay Multi-dimensional Scoring). %The first model is based on ``cardiffnlp/twitter-roberta-base-sentiment-latest''. The second one is termed as \texttt{DistilAEMS} because it is based on ``distilbert/distilbert-base-uncased-finetuned-sst-2-english''.  
Table \ref{tab1} provides the results on the the first model on the test dataset of ELLIPSE. The three criteria were taken to evaluate the model performance. The overall performance is beyond 0.8.  

\begin{table}[ht]
\centering
\caption{The performance of two models in the test dataset of ELLIPSE using the three criteria}
\scalebox{0.7}{
\begin{tabular}{lcccccc}
\toprule
\textbf{Dimension} & \multicolumn{2}{c}{\textbf{Precision}} & \multicolumn{2}{c}{\textbf{F1 Score}} & \multicolumn{2}{c}{\textbf{QWK}} \\
\cmidrule(lr){2-3} \cmidrule(lr){4-5} \cmidrule(lr){6-7}
 & \textbf{Ro-AEMS} & \textbf{Distil-AEMS} & \textbf{Ro-AEMS} & \textbf{Distil-AEMS} & \textbf{Ro-AEMS} & \textbf{Distil-AEMS} \\
\midrule
Cohesion    & 0.85 & 0.84 & 0.86 & 0.85 & 0.81 & 0.82 \\
Syntax      & 0.89 & 0.91 & 0.92 & 0.93 & 0.83 & 0.85 \\
Vocabulary  & 0.90 & 0.90 & 0.89 & 0.90 & 0.84 & 0.85 \\
Phraseology & 0.83 & 0.85 & 0.84 & 0.85 & 0.80 & 0.81 \\
Grammar     & 0.87 & 0.89 & 0.90 & 0.91 & 0.83 & 0.84 \\
Conventions & 0.91 & 0.82 & 0.92 & 0.92 & 0.84 & 0.85 \\
\bottomrule
\end{tabular}}
\label{tab1}
\end{table}

\subsection{Study 2}

Table \ref{tab2} provides the results on the the second model on the test dataset of IELTS. Note that there are five different dimensions in scoring. The same three criteria were taken to evaluate the model performance. The overall performance is also beyond 0.8.  Compared with Table \ref{tab1}, the model performance has slightly improved, probably because of the larger size of the training dataset and reduced number of scoring tasks (only five.). The information on hyperparameters and other setups during training is seen in \textbf{Appendix B}. 

\begin{table}[ht]
\centering
\caption{The performance of two models in the test dataset of IELTS using the three criteria}
\scalebox{0.68}{
\begin{tabular}{lcccccc}
\toprule
\textbf{Dimension} & \multicolumn{2}{c}{\textbf{Precision}} & \multicolumn{2}{c}{\textbf{F1 Score}} & \multicolumn{2}{c}{\textbf{QWK}} \\
\cmidrule(lr){2-3} \cmidrule(lr){4-5} \cmidrule(lr){6-7}
 & \textbf{Ro-AEMS} & \textbf{Distil-AEMS} & \textbf{Ro-AEMS} & \textbf{Distil-AEMS} & \textbf{Ro-AEMS} & \textbf{Distil-AEMS} \\
\midrule
Task Achievement    & 0.90 & 0.91 & 0.90 & 0.9 & 0.86 & 0.87 \\
Coherence and Cohesion      & 0.89 & 0.88 & 0.91 & 0.90 & 0.86 & 0.85 \\
Vocabulary  & 0.93 & 0.92 & 0.94 & 0.93 & 0.88 & 0.87 \\
%Phraseology & 0.83 & 0.85 & 0.84 & 0.85 & 0.80 & 0.82 \\
Grammar     & 0.91 & 0.89 & 0.89 & 0.87 & 0.84 & 0.85 \\
Overall & 0.89 & 0.88 & 0.91 & 0.89 & 0.85 & 0.86 \\
\bottomrule
\end{tabular}}
\label{tab2}
\end{table}

\section{Discussion}

Table \ref{tab3} lists the main achievements of various models on holistic scoring. Compared to these models, our model demonstrated better performance in holistic scoring based on the criterion of \texttt{QWK}. Additionally, our model can provide scores for multiple dimensions, with performance in these dimensions closely matching the overall scoring. 

\begin{table}[ht]
\centering
\caption{QWK performance of in the past AES models and closed-sourced models for overall scoring}
\begin{tabular}{lc}
\toprule
\textbf{Task} & \textbf{QWK Average (overall scoring)} \\
\midrule
Xie et al. (2022)  &  0.82 \\
Jiang et al. (2023) & 0.70 \\
ChatGPT  (Mansour et al., 2024)            & 0.31 \\
Llama    (Mansour et al., 2024)          & 0.30 \\
GPT-4V (Lee et al., 2023) & 0.43\\
\bottomrule
\end{tabular}
\label{tab3}
\end{table}

Furthermore, the two models exhibited similar performance across two types of test datasets, despite the differences in their dimensions. This cross-validation indicates that our models maintain stable performance across diverse datasets. It also demonstrates that our strategies are highly effective and robust in enhancing the AEMS system. The consistent results across varied datasets highlight the generalizability of our approach, ensuring reliable scoring under different conditions and validating the robustness of our model improvements.

\section*{Data Availability}

One AEMS model for the ELLIPSE corpus is available at: 
\burl {https://huggingface.co/Kevintu/Engessay_grading_ML}. The usage of the model is detailed in this huggingface repo. 

\bibliography{reference}
\bibliographystyle{plain}

\appendix

\section*{Appendix A: Incorporating contrastive learning for additional information}

Contrastive learning aims to make representations of similar inputs closer in the embedding space, while pushing representations of dissimilar inputs apart.

Let:
\begin{itemize}
    \item $\text{BERT}_{\text{cls}}(x)$ be the output representation of the [CLS] token for input $x$.
    \item $\text{BERT}_{\text{additional}}(r)$ be the representation of additional information $r$ (e.g., essay requirements, topic, type).
    \item $\mathbf{z}_i$ be the combined representation of the essay and its additional information.
\end{itemize}

%\section*{Combined Representation}

The combined representation $\mathbf{z}_i$ is obtained by concatenating the essay representation and the additional information representation:
\[
\mathbf{z}_i = \text{BERT}_{\text{cls}}(x_i) \oplus \text{BERT}_{\text{additional}}(r_i)
\]
where $\oplus$ denotes concatenation.

%\section*{Contrastive Loss}

For contrastive learning, we define the NT-Xent (Normalized Temperature-scaled Cross Entropy) loss. Let $(\mathbf{z}_i, \mathbf{z}_j)$ be a positive pair (e.g., essays with the same topic), and let $\tau$ be a temperature parameter. The NT-Xent loss for a positive pair is given by:
\[
\mathcal{L}_{\text{contrastive}}(\mathbf{z}_i, \mathbf{z}_j) = -\log \frac{\exp(\text{sim}(\mathbf{z}_i, \mathbf{z}_j) / \tau)}{\sum_{k=1}^{2N} \mathbb{1}_{[k \neq i]} \exp(\text{sim}(\mathbf{z}_i, \mathbf{z}_k) / \tau)}
\]
where $\text{sim}(\mathbf{z}_i, \mathbf{z}_j)$ denotes the cosine similarity between $\mathbf{z}_i$ and $\mathbf{z}_j$, and $N$ is the batch size.

%\section*{Combined Loss Function}

The overall loss function combines the classification loss, regression loss, and contrastive loss:
\[
\mathcal{L} = \mathcal{L}_{\text{CE}} + \lambda_1 \mathcal{L}_{\text{MSE}} + \lambda_2 \mathcal{L}_{\text{contrastive}}
\]

where:
\begin{itemize}
    \item $\mathcal{L}_{\text{CE}}$ is the cross-entropy loss for classification.
    \item $\mathcal{L}_{\text{MSE}}$ is the mean squared error loss for regression.
    \item $\mathcal{L}_{\text{contrastive}}$ is the contrastive loss.
    \item $\lambda_1$ and $\lambda_2$ are hyperparameters that balance the contributions of the different loss components.
\end{itemize}

By optimizing this combined loss function, the model can better understand and utilize additional information such as essay requirements, topics, and types, leading to improved automatic essay scoring.

\section*{Appendix B: Hyperparaters and setups in training}
Tables \ref{para1} and \ref{para2} provides the information on some of the hyperparameters in training based on the two existing models. The full hyperparameters are given on request.

\begin{table}[h!]
\centering
\caption{Training configuration parameters in RoBERT}
%\label{tab:training_params}
\begin{tabular}{|l|l|}
\hline
\textbf{Parameter} & \textbf{Value} \\ \hline
num\_train\_epochs & 28 \\ \hline
%per\_device\_train\_batch\_size & 16 \\ \hline
per\_device\_eval\_batch\_size & 16 \\ \hline
warmup\_steps & 500 \\ \hline
%weight\_decay & 0.01 \\ \hline
%logging\_dir & ./logs \\ \hline
logging\_steps & 10 \\ \hline
evaluation\_strategy & epoch \\ \hline
%save\_strategy & epoch \\ \hline
%load\_best\_model\_at\_end & True \\ \hline
%metric\_for\_best\_model & eval\_loss \\ \hline
learning\_rate & 2e-5 \\ \hline
%warmup\_ratio & 0.1 \\ \hline
%contrastive\_learning\_temperature & 0.07 \\ \hline
contrastive\_learning\_batch\_size & 128 \\ \hline
%loss\_weight\_classification & 1.0 \\ \hline
%loss\_weight\_regression & 0.5 \\ \hline
%loss\_weight\_contrastive & 0.5 \\ \hline
\end{tabular}
\label{para1}
\end{table}

\begin{table}[h!]
\centering
\caption{Training configuration parameters in DistilBERT}
%\label{tab:training_params}
\begin{tabular}{|l|l|}
\hline
\textbf{Parameter} & \textbf{Value} \\ \hline
num\_train\_epochs & 22 \\ \hline
%per\_device\_train\_batch\_size & 16 \\ \hline
per\_device\_eval\_batch\_size & 64 \\ \hline
warmup\_steps & 500 \\ \hline
%weight\_decay & 0.01 \\ \hline
logging\_steps & 10 \\ \hline
evaluation\_strategy & epoch \\ \hline
%save\_strategy & epoch \\ \hline
%load\_best\_model\_at\_end & True \\ \hline
%metric\_for\_best\_model & eval\_loss \\ \hline
learning\_rate & 2e-5 \\ \hline
%contrastive\_learning\_temperature & 0.08 \\ \hline
contrastive\_learning\_batch\_size & 130 \\ \hline
%loss\_weight\_classification & 1.0 \\ \hline
%loss\_weight\_regression & 0.6 \\ \hline
%loss\_weight\_contrastive & 0.6 \\ \hline
\end{tabular}
\label{para2}
\end{table}

\end{document}